# Inherited Weights: Measuring the Tokenization Premium and the Sovereignty Cost of Language Models for the Global South

Paulo Câmara
LUA Vision Tecnologia, São Paulo, Brazil · ORCID 0009-0007-7607-4179 · paulo@lua.vision

**Abstract.** Most language models deployed outside the United States and China are fine-tuned descendants of a small number of pretrained families. We argue that fine-tuning leaves the decisive properties of a model inherited from its base: vocabulary, knowledge, cultural alignment, memorized data and license. We contribute three things. (i) An updated, reproducible measurement of the tokenization premium across 31 languages and 7 tokenizers released in 2023–2024, on 27 articles of the Universal Declaration of Human Rights present in every language, with explicit correction for unknown-token collapse. Premiums reach 10× for Amharic in older tokenizers and remain at 6.8× in the newest commercial one; Indigenous South American languages pay between 1.6× and 2.7× under global tokenizers. A vocabulary built from Portuguese lowers the premium for 11 of 13 Latin-script Global South languages relative to every global tokenizer tested, by about 24% for Guarani and Quechua. (ii) An eight-test black-box sovereignty audit that a buyer can run without access to weights, each test anchored in a published result. (iii) A regime argument: under energy and liability constraints, accuracy per joule, calibration and selective risk should replace raw accuracy as design objectives, and we give the metrics in a form that prevents abstention from inflating scores. We close with preliminary internal measurements from a proprietary architecture built around these objectives, reported with their limitations.

**Keywords:** tokenization; multilingual fairness; Global South; fine-tuning; calibration; selective prediction; energy efficiency; AI sovereignty.

## 1 Introduction

"National" and "sovereign" language models are announced every month. In most cases the artifact behind the announcement is a model pretrained elsewhere, adapted with a comparatively small amount of local data. The label promises that the model was conceived, built and governed locally. The weights usually say otherwise.

This matters for a technical reason, not only a political one. A language model is a function whose behavior is fixed, to first order, by its pretrained parameters: the tokenizer decides how text is cut and priced, the embeddings and weights encode what the model knows and how it associates concepts, and the pretraining corpus determines what it memorized. Fine-tuning reshapes style and format with remarkably little data [1], but it does not reliably add knowledge; new facts introduced by fine-tuning are learned slowly and increase hallucination roughly linearly as they are absorbed [2]. The opinions a model reflects track the populations that dominate its pretraining data [3] [4], and behaviors implanted in pretraining can survive safety training [5]. Licenses travel with derivatives: the Llama 3.1 Community License, for example, requires that distributed derivatives carry "Llama" at the beginning of their name [6].

The first cost a user in the Global South meets is the tokenizer. Commercial models are priced per token, and the same content costs more tokens in some languages than in others [7] [8]. We update that measurement with 2024 tokenizers, extend it to African and Indigenous South American languages, and report a result we did not expect: a vocabulary built from Portuguese is cheaper than every global tokenizer tested for most Latin-script languages of the Global South.

Section 2 fixes vocabulary. Section 3 presents the measurement. Section 4 gives the audit. Section 5 states the regime argument and the metrics. Section 6 reports preliminary measurements of an architecture designed around those metrics. Section 7 lists limitations.

## 2 What is inherited

We use "model" for the tuple (tokenizer, embeddings, weights, architecture, training recipe). A **family** is a set of models sharing a pretrained ancestor. Fine-tuning, including instruction tuning and preference optimization, modifies weights starting from that ancestor and, in common practice, changes a small fraction of the information content of the parameters.

We distinguish five inheritances. **What it knows**: knowledge is acquired in pretraining; fine-tuning mostly teaches format [1] and adding new facts through it raises hallucination [2]. **How it reads**: the tokenizer is frozen with the embeddings, so changing it requires re-learning the input layer. **How it thinks**: associations, cultural defaults and opinion distributions come from the pretraining mix [3] [4]. **What cannot be seen**: memorized training data is extractable [9], poisoning at web scale is practical [10], and backdoors can persist [5]. **Whose it is**: the license of the base binds the derivative [6].

None of these is changed by the label on the derivative. Each can be measured, at least partly, from outside.

## 3 Measuring the tokenization premium

### 3.1 Data and tokenizers

We use the Universal Declaration of Human Rights as distributed by the UDHR in Unicode project [11], which provides normalized XML for hundreds of languages. We keep the 27 articles present in all 31 selected languages (articles 1, 2, 4, 5, 7, 8 and 10–30), covering European, Asian and African languages, Indigenous languages of the Americas and two creoles. The declaration is a parallel, legally reviewed text: every version says the same thing, which is the property a price comparison needs.

We tokenize with seven tokenizers: five from open models, Pythia [12], Mistral 7B, Qwen 2.5, DeepSeek-V3 and the Brazilian Tucano [13]; and two public commercial encodings, cl100k_base (2023) and o200k_base (2024), from the tiktoken library. Open tokenizers are loaded from their published `tokenizer.json` files with the HuggingFace `tokenizers` library.

### 3.2 Metric

For language $\ell$ and tokenizer $\tau$, the premium is the ratio of token counts on the same content:

$$P_\tau(\ell) = \frac{N_\tau(\ell)}{N_\tau(\text{en})} \tag{1}$$

where $N_\tau$ is the total number of tokens over the 27 articles. $P = 1$ means parity with English; $P = 2$ means the same content costs twice as many tokens and therefore, under per-token pricing, twice as much money and twice as much context.

**Unknown-token collapse.** A tokenizer that does not cover a script may map large spans to a single unknown token. This produces a spuriously low count. Tucano's vocabulary replaces up to half of the text with its unknown token in non-Latin scripts; we report those cells as n/a rather than as low premiums. Byte-level tokenizers (all others here) never emit unknown tokens.

**Robustness.** Two independent Brazilian Portuguese translations of the declaration (UNICEF Brazil and UDHR in Unicode) give premiums of 1.22 and 1.15 under the same tokenizer, so the result does not hinge on one translation.

### 3.3 Results

| Language | Pythia 2023 | Mistral 2023 | cl100k 2023 | Qwen 2.5 2024 | DeepSeek 2024 | o200k 2024 | Tucano 2024 |
|---|---|---|---|---|---|---|---|
| European | | | | | | | |
| English | 1.00 | 1.00 | 1.00 | 1.00 | 1.00 | 1.00 | 1.00 |
| Portuguese (Brazil) | 1.67 | 1.62 | 1.47 | 1.45 | 1.35 | 1.15 | **0.64** |
| Spanish | 1.68 | 1.59 | 1.47 | 1.45 | 1.41 | 1.23 | **1.03** |
| French | 1.77 | 1.63 | 1.58 | 1.56 | 1.52 | 1.32 | **1.28** |
| German | 1.98 | 1.72 | 1.67 | 1.65 | 1.54 | 1.28 | 1.50 |
| Russian | 2.88 | 2.00 | 2.61 | 1.74 | 1.54 | 1.37 | 3.85 |
| Asia | | | | | | | |
| Chinese (simpl.) | 1.85 | 1.52 | 1.70 | 0.90 | 0.84 | 1.13 | n/a |
| Japanese | 2.40 | 2.40 | 2.58 | 1.54 | 1.45 | 1.91 | n/a |
| Korean | 3.64 | 2.46 | 2.43 | 1.46 | 1.55 | 1.36 | n/a |
| Hindi | 5.55 | 5.54 | 5.41 | 5.15 | 2.97 | 1.53 | n/a |
| Bengali | 6.90 | 5.38 | 6.11 | 5.32 | 2.09 | 1.69 | n/a |
| Tamil | 9.71 | 7.21 | 9.55 | 7.52 | 3.63 | 2.31 | n/a |
| Vietnamese | 3.80 | 4.33 | 4.45 | 1.47 | 4.07 | 3.56 | n/a |
| Indonesian | 2.41 | 2.43 | 1.92 | 1.92 | 1.56 | 1.47 | 1.61 |
| Arabic | 2.69 | 3.34 | 2.72 | 1.39 | 1.39 | 1.19 | n/a |
| Africa | | | | | | | |
| Swahili | 2.08 | 2.02 | 1.98 | 1.98 | 1.97 | 1.42 | 1.44 |
| Yorùbá | 5.46 | 4.75 | 4.63 | 4.24 | 4.49 | 3.14 | n/a |
| Hausa | 3.15 | 3.04 | 2.96 | 2.86 | 2.85 | 2.15 | n/a |
| Igbo | 3.24 | 2.70 | 2.70 | 2.60 | 2.60 | 1.75 | n/a |
| Amharic | 10.02 | 8.30 | 9.99 | 3.83 | 7.69 | 6.76 | n/a |
| Zulu | 2.22 | 2.17 | 2.07 | 2.07 | 2.08 | 1.62 | **1.47** |
| Somali | 2.61 | 2.55 | 2.45 | 2.44 | 2.48 | 1.82 | **1.75** |
| Wolof | 2.10 | 2.05 | 1.98 | 1.97 | 2.00 | 1.77 | **1.40** |
| Lingala | 2.07 | 1.99 | 1.87 | 1.87 | 1.89 | 1.66 | **1.44** |
| Kinyarwanda | 1.96 | 1.94 | 1.84 | 1.84 | 1.83 | 1.47 | **1.31** |
| Nigerian Pidgin | 2.21 | 2.11 | 2.16 | 2.16 | 2.18 | 2.13 | **1.74** |
| Indigenous Americas and creoles | | | | | | | |
| Guarani | 2.67 | 2.49 | 2.53 | 2.39 | 2.37 | 2.11 | **1.61** |
| Quechua | 1.84 | 1.81 | 1.76 | 1.76 | 1.77 | 1.59 | **1.21** |
| Aymara | 2.08 | 2.07 | 1.99 | 1.98 | 1.97 | 1.79 | **1.41** |
| Haitian Creole | 2.13 | 2.04 | 1.99 | 1.99 | 1.92 | 1.58 | **1.45** |
| Tok Pisin | 1.94 | 1.85 | 1.84 | 1.84 | 1.81 | 1.53 | **1.27** |

Table 1: Tokenization premium relative to English (English = 1.00) on 27 articles of the UDHR. Bold: Tucano below every global tokenizer. n/a: unknown-token collapse.

Four findings stand out in Table 1.

**The premium grows with distance from English and European languages.** In older tokenizers Amharic costs 10.0× English, Tamil 9.7×, Bengali 6.9× and Yorùbá 5.5×.

**Indigenous South American languages pay about double.** Guarani, Quechua and Aymara cost between 1.6× and 2.7× English under global tokenizers.

**Recent improvement is real and uneven.** The 2024 commercial encoding lowers almost every premium, yet Amharic remains at 6.8×, Vietnamese at 3.6× and Yorùbá at 3.1×. The languages that were furthest away remain furthest away.

**A vocabulary built outside English helps other Southern languages.** Tucano's Portuguese-built vocabulary is cheaper than all six global tokenizers for 11 of 13 Latin-script Global South languages in our set (the exceptions are Swahili and Indonesian). For Guarani and Quechua it is about 24% below the best global tokenizer (1.61 vs. 2.11 and 1.21 vs. 1.59). Part of this effect plausibly comes from Iberian-derived orthography shared with Portuguese; separating orthographic overlap from vocabulary design requires a controlled comparison, for instance against a Spanish-built vocabulary and on FLORES-200, which we leave to the next version.

The same vocabulary fails on non-Latin scripts. A local vocabulary solves the local problem and creates others; the finding argues for vocabularies designed from the languages of use, not for any single existing one.

### 3.4 Relation to prior work

Rust et al. showed that tokenizer quality explains part of the gap between monolingual and multilingual models [14]. Petrov et al. quantified premiums of up to 15× across languages for 2023-era tokenizers [7], and Ahia et al. connected the premium to the price and utility of commercial APIs [8]. Two recent studies are the closest to ours. Churchill and Skiena analyze the tokenizers of ten popular models and propose post-hoc vocabulary additions that reduce the premium for 12 low-resource languages [23]. Roy et al. audit three tokenizers on a Python tutoring corpus in five languages and report, for o200k, 1.56× for Bengali and 2.37× for Yorùbá [24], in the same direction as our 1.69× and 3.14× on a different text. We add a legally parallel text in 31 languages, African and Indigenous South American coverage, seven tokenizers from 2023–2024, explicit unknown-token correction, and the observation about a Portuguese-built vocabulary.

## 4 An eight-test sovereignty audit

A buyer rarely receives weights. The following tests run against an API or a chat interface. Each is anchored in a published result and each measures one inheritance from Section 2.

| # | Test | Procedure | Grounding |
|---|---|---|---|
| 1 | Provenance | Ask for base family, license and training cutoff; compare with public configuration files and license terms. | [6] |
| 2 | Tokenization premium | Count tokens for the same text in the buyer's language and in English; compute $P$. | [7] [8] |
| 3 | Data egress | Inspect where requests terminate, retention terms and whether inputs train future models. | contract |
| 4 | Calibration and abstention | Ask questions with known answers and questions outside the model's knowledge; record stated confidence and refusals. | [15] [16] [17] |
| 5 | Memorization | Prompt with the opening of well-known texts and measure verbatim continuation. | [9] |
| 6 | Indirect injection | Embed an instruction in a document the system reads; check whether it is obeyed. | [18] [19] |
| 7 | Instruction leakage | Ask, in several phrasings, for "the text before this conversation". | [20] |
| 8 | Consistency | Repeat the same business question in ten fresh sessions; count distinct answers. | — |

Table 2: Black-box sovereignty audit. All tests can be run without access to weights.

A practical caveat applies to tests 4, 6 and 7. Much of the protection users observe in large assistants lives outside the model, in input and output filters, classifiers and supervising agents. The same base model reached through a different endpoint can behave differently. The audit therefore measures the system that is being bought, and should be repeated if the access path changes.

## 5 The regime argument

### 5.1 Why raw accuracy is the wrong objective under constraints

Levy and Baxter showed in 1996 that the neural code maximizing information capacity is not the one maximizing information per unit of energy once the cost of activity enters the objective [21]. The brain operates near the second optimum, at roughly 20 W. Current language models are optimized for the first. In settings where energy is scarce and errors carry liability, such as credit, health triage and public services, the relevant question is not how often the model is right but how much each correct answer costs and whether the model knows when it is wrong.

### 5.2 Metrics that abstention cannot inflate

Let a model answer $n$ items with confidence $c_i \in [0, 1]$ and correctness $y_i \in \{0, 1\}$. Calibration error is computed over **all** items, answered or not:

$$\mathrm{ECE} = \sum_{b=1}^{B} \frac{|B_b|}{n} \cdot |\mathrm{acc}(B_b) - \mathrm{conf}(B_b)| \tag{2}$$

Selective performance is summarized by the risk–coverage curve: sort items by decreasing confidence, and for coverage $k/n$ record the error rate of the top $k$. The area under this curve (AURC) and its excess over an oracle ordering (E-AURC) reward models whose confidence ranks their own errors correctly [22]. We also report the AUROC of confidence for separating in-domain from out-of-domain items. Finally,

$$\text{accuracy per joule} = \frac{\sum_i y_i}{E} \tag{3}$$

where $E$ is the measured energy at the wall for the full evaluation. Reporting accuracy only on answered items, without coverage and without ECE on all items, lets a model raise its score by refusing; the set above closes that loophole.

## 6 Preliminary measurements from a reference design

LUA Vision is building an architecture around the objectives of Section 5. The design is proprietary and is not named or described here. We report preliminary internal measurements so that the claims in this paper about what is possible are tied to numbers, and we list their limitations explicitly. On an internal calibration benchmark mixing in-domain and out-of-domain questions, the model was trained on a laptop, with the base stage completing in about 13 s and responses taking about 0.25 s each, without a data-center accelerator. Results:

- ECE computed over all items: 0.122.
- When the model's confidence was at least 0.6, accuracy was 94.3% (35 answered items); at confidence at least 0.7, accuracy was 100% (23 items).
- Mean confidence fell from 0.297 on in-domain items to 0.164 on out-of-domain items; AUROC for in vs. out of domain: 0.664.
- E-AURC: 0.012.

These figures are preliminary. The benchmark is internal and small, the runs were executed by the author's team, and the results have not been independently reproduced. A second probe on a different task was degenerate (AUROC 0.53 with a single confident answer out of 20) and is reported here for completeness. Evaluation code implementing the metrics of Section 5 is available from the author.

Two further properties are design targets of the LUA tokenizer and pricing model rather than measurements: a vocabulary in which a word maps to one token across the languages of use, which would drive the premium $P$ toward 1, and billing units other than usage — a licence, installed capacity, or a task delivered and verified — the last of which is only viable for a model whose confidence predicts its own correctness.

## 7 Limitations

The premium is measured on one parallel text; legal register is not everyday language, and FLORES-200 should confirm the ranking. Token counts approximate cost but prices also depend on tier, caching and currency. The Portuguese-vocabulary effect may be partly orthographic, as discussed. The audit tests are qualitative instruments, not a certification. The measurements in Section 6 are internal, small-sample and preliminary. The author founded the company that builds the reference design, which is a conflict of interest declared below.

## 8 Conclusion

For the Global South, the question that matters about a language model is not where it was fine-tuned but whose weights it runs, how it reads the local language, what it does when it does not know, and where the data goes. The first of these is now measurable with public tools: on the same text, the newest commercial tokenizer still charges 6.8 times more for Amharic than for English, and a vocabulary built from Portuguese already beats every global tokenizer for most Latin-script languages of the South. The rest can be tested from outside with eight procedures. Designing for accuracy per joule, calibration and selective risk is a different engineering target from the one the market optimizes today, and we believe it is the right one where energy is scarce and errors have victims.

**Conflict of interest.** The author is the founder of LUA Vision Tecnologia, which develops language models and the reference design described in Section 6.

**Data and code availability.** Token counts and the measurement script are deposited with this preprint. Texts are from the UDHR in Unicode project; tokenizers are public.

**Companion essay.** A longer Portuguese essay by the author, "O peso que você não escolheu", presents the same argument for a general audience at lua.vision/pesquisa/o-peso-que-voce-nao-escolheu.